%% file: main.tex
\documentclass{article}

\usepackage{amsmath}

\usepackage[dvipsnames]{xcolor}

\usepackage{float}

\usepackage{graphicx} 
\graphicspath{{images/}}
\DeclareGraphicsExtensions{.pdf,.jpeg,.png}

\usepackage[caption=false]{subfig}

\usepackage{multicol}
\usepackage{multirow}

\usepackage{url}

\usepackage{comment}
\usepackage{verbatim}

\usepackage{booktabs}
\usepackage{array}

\usepackage{hyperref}

\usepackage[accepted]{icml2018}

\icmltitlerunning{Leveraging Random Label Memorization for Unsupervised Pre-Training}

\begin{document}

\twocolumn[
\icmltitle{Leveraging Random Label Memorization for Unsupervised Pre-Training}



\icmlsetsymbol{equal}{*}

\begin{icmlauthorlist}
\icmlauthor{Vinaychandran~Pondenkandath}{equal,unifr}
\icmlauthor{Michele~Alberti}{equal,unifr}
\icmlauthor{Sammer~Puran}{unifr}
\icmlauthor{Rolf~Ingold}{unifr}
\icmlauthor{Marcus~Liwicki}{unifr,lulea}
\end{icmlauthorlist}

\icmlaffiliation{unifr}{Document Image and Voice Analysis Group (DIVA), University of Fribourg, Fribourg, Switzerland}
\icmlaffiliation{lulea}{Machine Learning Group, Lule{\aa} University of Technology, Lule{\aa}, Sweden}

\icmlcorrespondingauthor{Michele~Alberti}{michele.alberti@unifr.ch}
\icmlcorrespondingauthor{Vinaychandran~Pondenkandath}{vinaychandran.pondenkandath@unifr.ch}

\icmlkeywords{Machine Learning, ICML}

\vskip 0.3in
]



\printAffiliationsAndNotice{\icmlEqualContribution} 

\input{sections/abstract}

\input{sections/introduction.tex}

\input{sections/experimental_setup.tex}

\input{sections/results.tex}

\input{sections/conclusion.tex}

\section*{Acknowledgment}
The work presented in this paper has been partially supported by the HisDoc~III project funded by the Swiss National Science Foundation with the grant number $205120$\textunderscore$169618$.


\bibliography{biblio}
\bibliographystyle{icml2018}

\end{document}

%% file: sections/abstract.tex
\begin{abstract}

We present a novel approach to leverage large unlabeled datasets by pre-training state-of-the-art deep neural networks on randomly-labeled datasets. 
Specifically, we train the neural networks to memorize arbitrary labels for all the samples in a dataset and use these pre-trained networks as a starting point for regular supervised learning.
Our assumption is that the `memorization infrastructure' learned by the network during the random-label training proves to be beneficial for the conventional supervised learning as well.  
We test the effectiveness of our pre-training on several video action recognition datasets (HMDB51, UCF101, Kinetics) by comparing the results of the same network with and without the random label pre-training. 
Our approach yields an improvement --- ranging from 1.5\,\% on UCF-101 to 5\,\% on Kinetics --- in classification accuracy, which calls for further research in this direction.

\end{abstract}

%% file: sections/introduction.tex
\section{Introduction}
\label{toc:introduction}

\begin{figure}[!t]
    \includegraphics[width=\columnwidth]{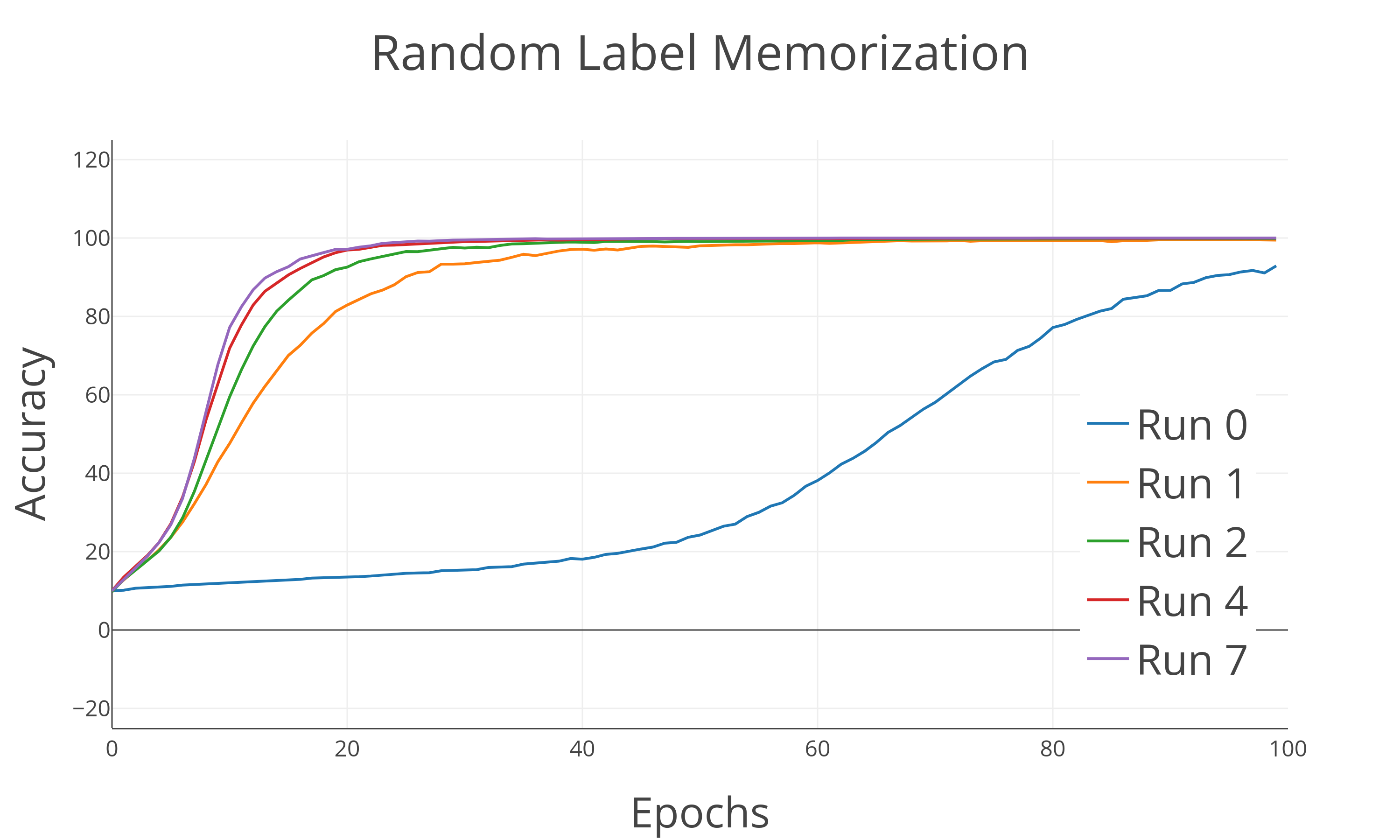}
    \caption{Training accuracy of the same network while sequentially re-shuffling the set of labels, hence forcing the network to start over with the memorization process.The first time, from random weights initialization, it learns very slowly. Later on, subsequent memorization processes are much faster than the initial one, thus indicating that the network is in fact building up a memorization infrastructure of some sort.}
    \label{fig:retrain}
\end{figure}

The success of many deep learning systems relies on supervised learning with a very large amount of labeled data. 
However, labeling data is an expensive process both in terms of time and money.
Even with the most advanced crowd sourcing techniques, it requires a significant amount of effort and yet the results are not guaranteed to be accurate.

In contrast, the mere collection of large amounts of data is fairly trivial as it is easily available online, e.g., several hundred hours of videos are uploaded to YouTube every minute~\cite{youtubestats}.
For this reason, developing improved unsupervised learning methods are of particular interest, as they can leverage large amounts of unlabeled data and extract meaningful information without supervision. 
However, developing effective methods to do this is not at all trivial.
    
After the introductions of greedy layer-wise training~\cite{hinton2006fast}, there have been numerous previous attempts which use simple~\cite{Jain2010} and advanced~\cite{dundar2015convolutional} clustering techniques, introduce surrogate classes~\cite{Dosovitskiy2014}, use Generative Adversarial Networks (GAN) \cite{Radford2015}, or use Auto-Encoders~\cite{Masci2011, Baldi2012, Bengio2013, Yang2015, Zhao2015}.

In particular, using unsupervised techniques as pre-training for a later classification task is a long known approach~\cite{Bengio2007, Erhan2009, Glorot2010}. 
However, despite the evident advantages of using unsupervised pre-training~\cite{erhan2010does}, common machine learning experience and recent work suggests that training for reconstruction first, and for classification later might not be the best idea in all cases~\cite{alberti2017}.
This is due to the inherently different nature of the two tasks (reconstruction and classification) which leads deep neural networks to learn different features for solving them.

In this paper we choose to approach unsupervised learning from a different direction: instead of pre-training for reconstruction, we pre-train the network to memorize randomly-assigned labels for all the samples in a dataset. 
This allows us to train a classification task on datasets that do not have any labels.

This work is inspired by the recent intriguing findings about the capacity of deep neural networks to memorize the training set~\cite{Zhang2016} and the possibility to measure the intrinsic dimension of the objective landscape~\cite{Li2018}.
The former rigorously show that deep neural networks are capable of overfitting to a training set even when there is no correlation between the labels and data, i.e., the training labels are shuffled.
This suggests that some type of features have to be learned by the network to succeed in this task, although they could be arbitrarily specialized to identify some sort of noise or bias in the input images~\cite{alberti2018tampering}.
In the latter, however, the authors observe that there is a generalization from one part of the training set to another, even though the label are just random. 
Furthermore, they make the hypothesis that training on some random labels forces the network to setup kind of a base infrastructure for memorization, and that this infrastructure can then be used to make further memorization more efficient\footnote{This has been extracted from \url{https://www.youtube.com/watch?v=uSZWeRADTFI}}. 
For these reasons we believe that pre-training with random labels might lead the network to learn useful features that can be used as a starting point (we are only pre-training after all) for further supervised training.  

\subsection*{Contribution}

In this work we introduce a novel approach for performing unsupervised pre-training in the form of training for classification with random labels.
Our preliminary experiments show that is it possible to learn useful representations by leveraging large amount of unlabeled data which calls for further research in this direction. 

\begin{figure*}[!t]
    \subfloat[]%
    {\includegraphics[width=.47\textwidth]{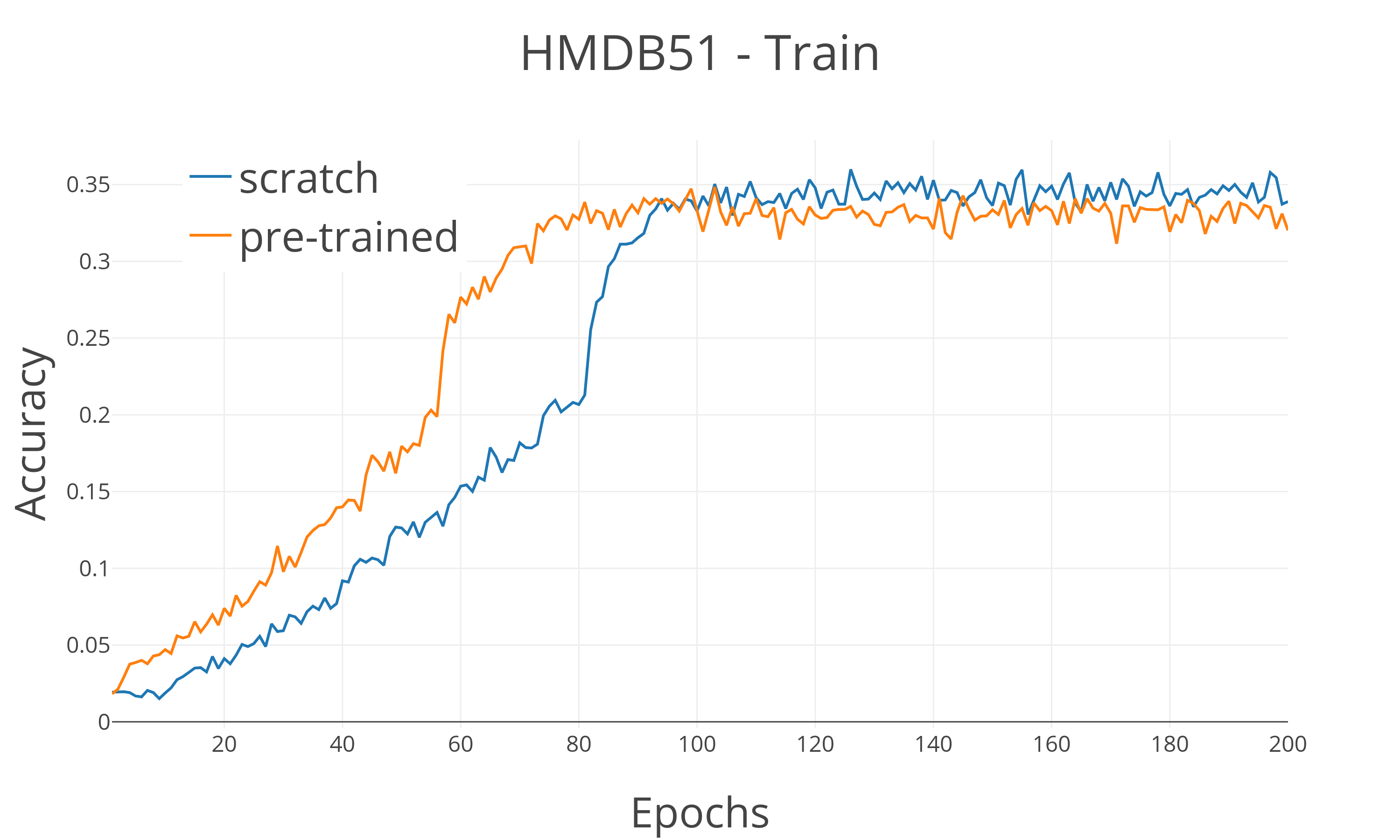}%
    \label{subfig:hmdb_train}}
    \hfil
    \subfloat[]%
    {\includegraphics[width=.47\textwidth]{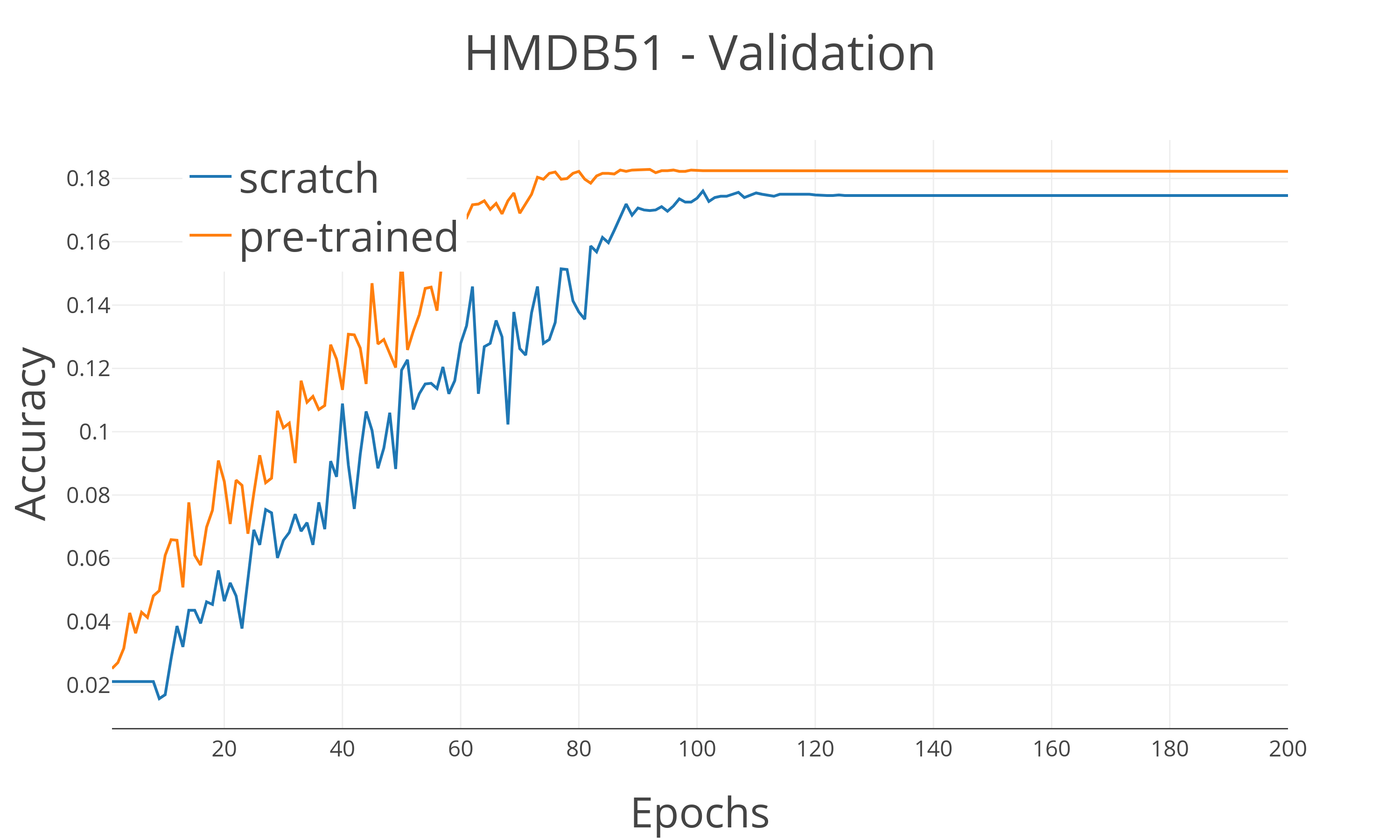}%
    \label{subfig:hmdb_val}}
    
    \vfil 
    
    \subfloat[]%
    {\includegraphics[width=.47\textwidth]{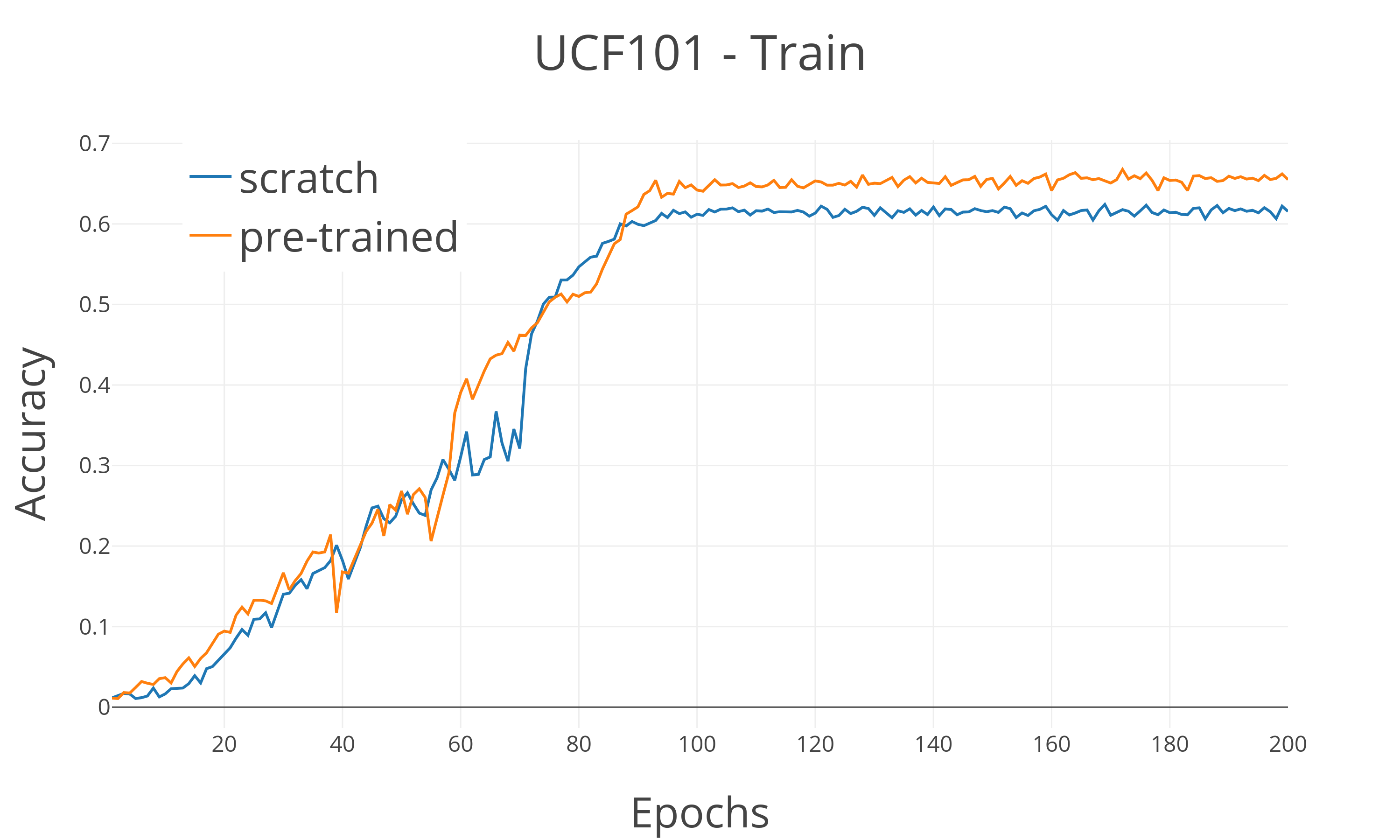}%
    \label{subfig:ucf_train}}
    \hfil
    \subfloat[]%
    {\includegraphics[width=.47\textwidth]{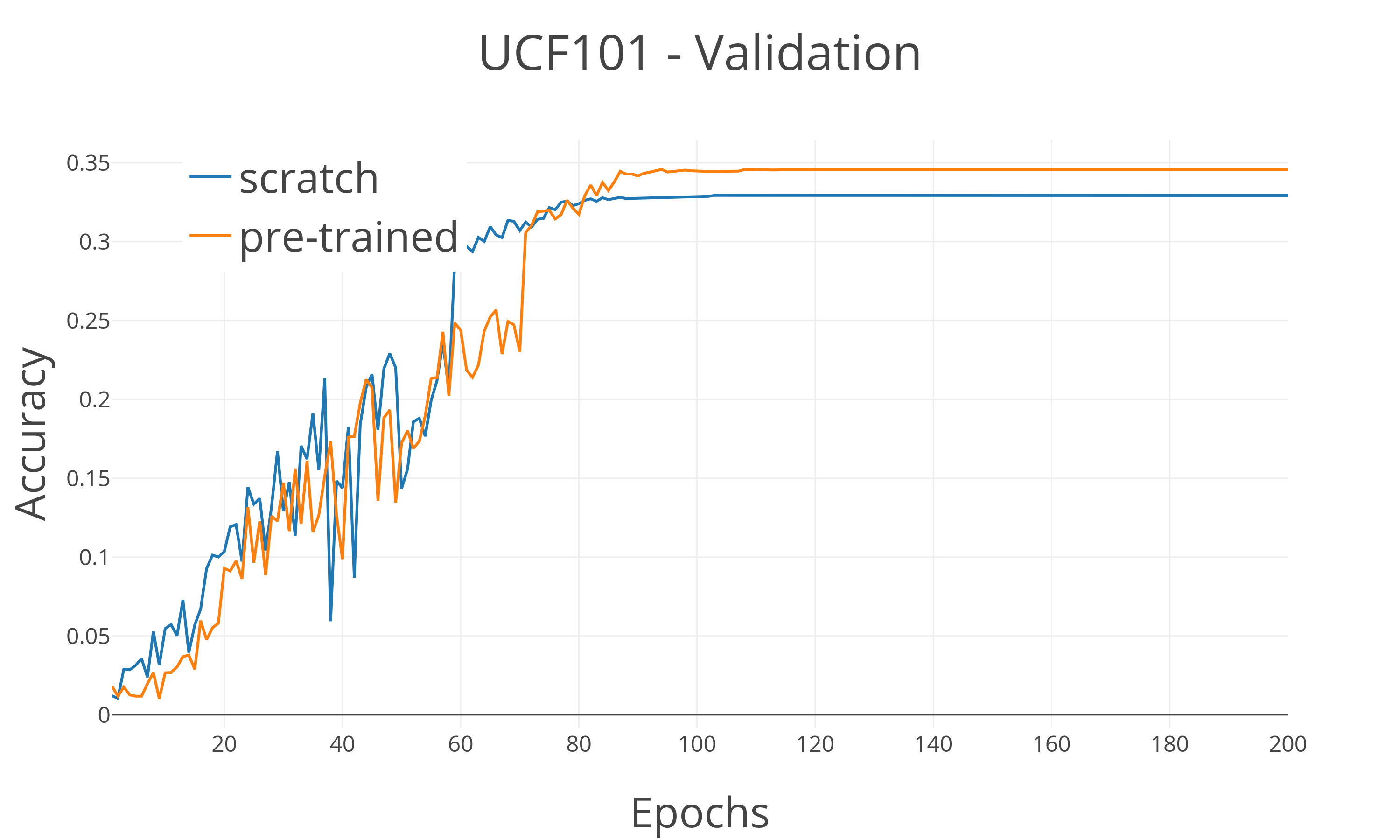}%
    \label{subfig:ucf_val}}
    
    \caption{Accuracy curves on the HMDB51~(\textit{a.} and \textit{b.}) and UCF-101~(\textit{c.} and \textit{d.}) datasets, where C3D networks have been either randomly initalized (\textcolor{CornflowerBlue}{blue line}) or cross pre-trained on a randomly-labeled variant of the complementary dataset (\textcolor{YellowOrange}{orange line}). For example, in \textit{a.}/\textit{b.} the orange line refers to a C3D network that was pre-trained on a randomly-labeled variant of the UCF-101 dataset. In these plots, we can see that the pre-training on random labels proves to be beneficial with an improvement in the accuracy of $0.8\%$ for HMDB51 and $1.7\%$ for the UCF-101 dataset. }
    \label{fig:hmdb_ucf}
\end{figure*}

\begin{figure*}[!t]
    \subfloat[]%
    {\includegraphics[width=.47\textwidth]{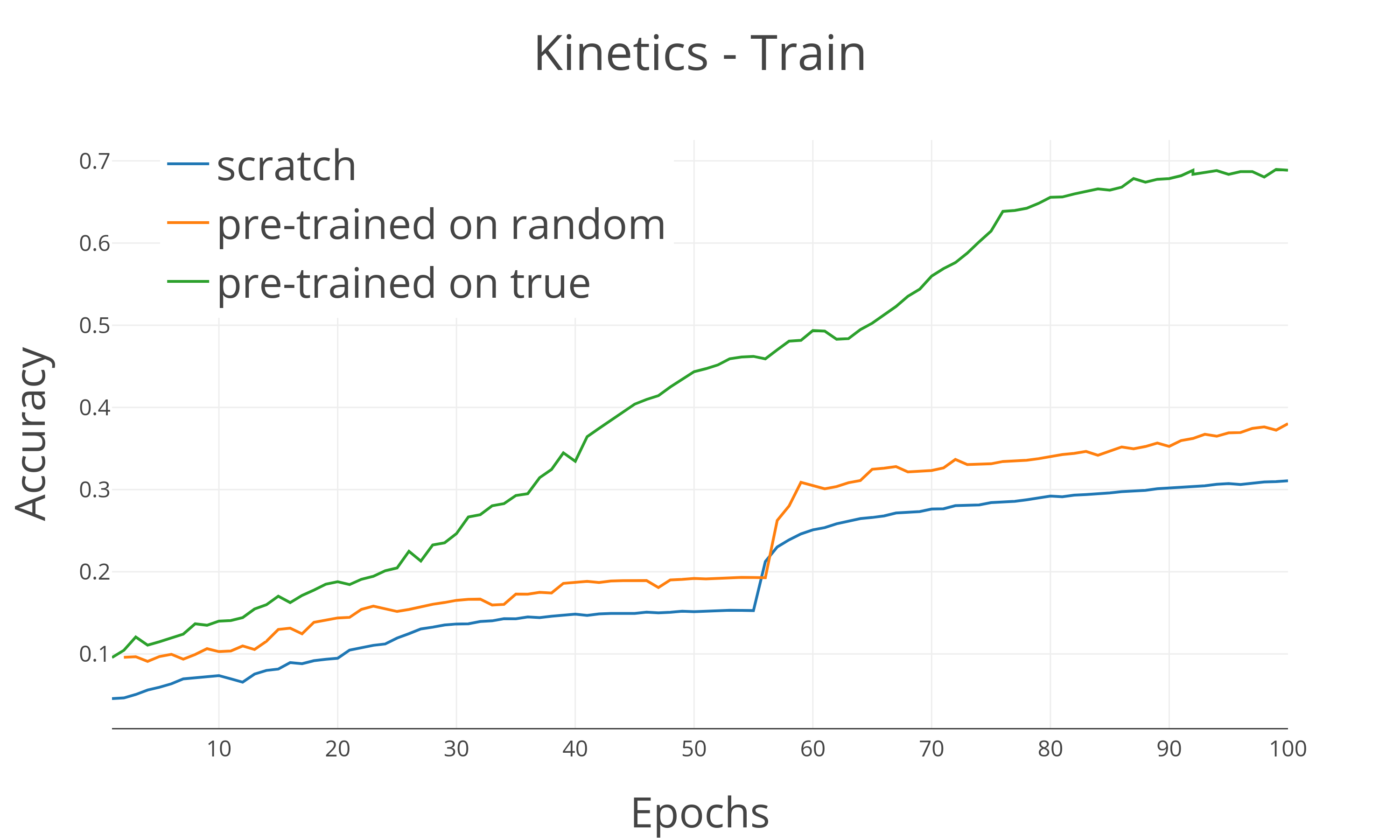}%
    \label{subfig:C3D_kinetics_train}}
    \hfil
    \subfloat[]%
    {\includegraphics[width=.47\textwidth]{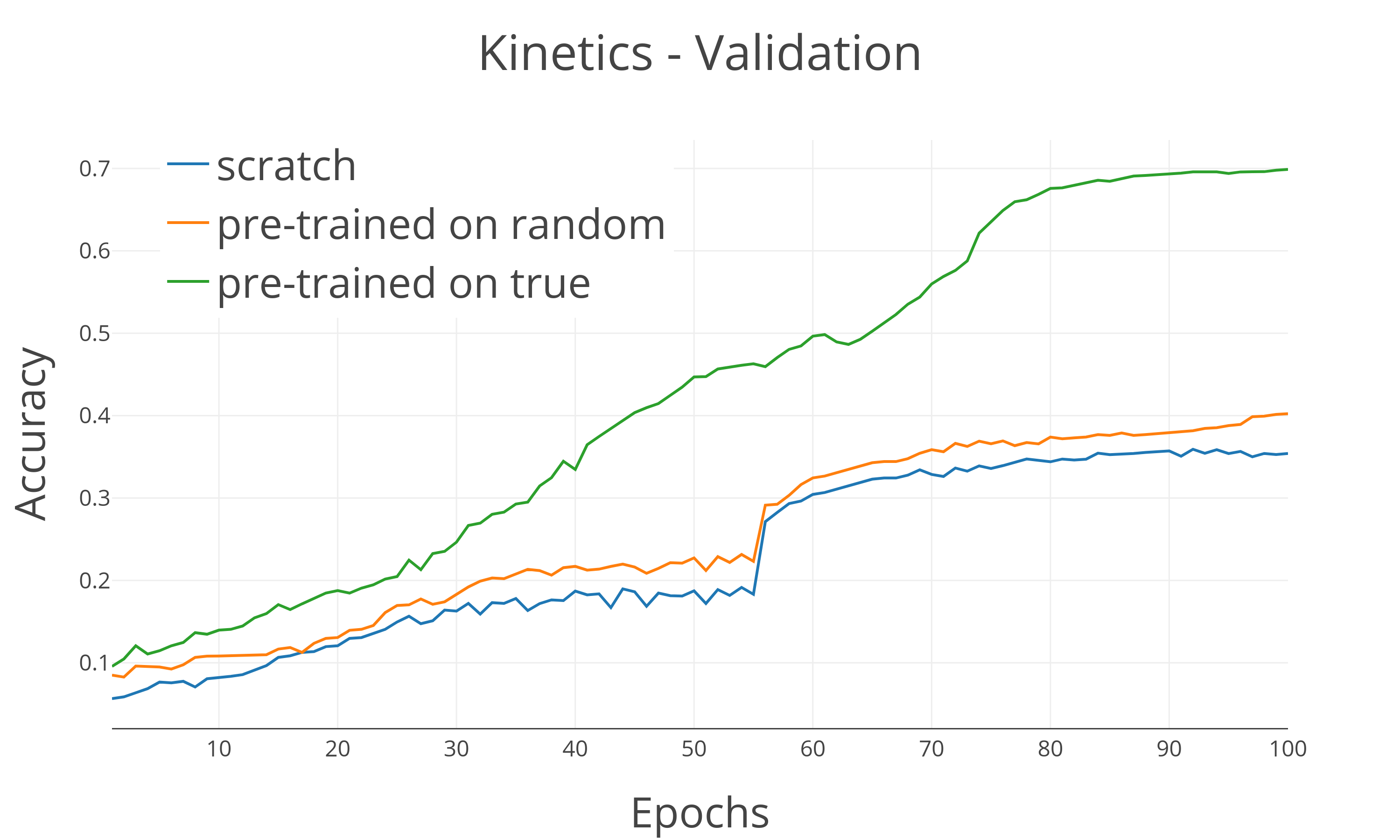}%
    \label{subfig:C3D_kinetics_val}}
    \caption{Accuracy curves on the Kinetics dataset comparing the performance of C3D networks that have been: randomly initialized (\textcolor{CornflowerBlue}{blue line}), pre-trained on a randomly-labeled subset of the Sports-1M dataset (\textcolor{YellowOrange}{orange line}) or pre-trained on the Sports-1M dataset with the correct labels (\textcolor{ForestGreen}{green line}). The blue line shows the baseline performance of C3D, the green line shows the upper bound of transfer learning from the complete Sports-1M dataset and the orange line shows the improved performance from our unsupervised pre-training method. It's important to note that only a subset ($40\%$) of the Sports-1M dataset was used for the random-label pre-training due to computational constraints. }
    \label{fig:C3D_kinetics}
\end{figure*}

%% file: sections/experimental_setup.tex
\section{Experimental Setting}
\label{toc:experimental_setting}

In this section we explain the experimental setup, i.e., the task performed, the dataset and the model used as well as the training procedure such that our experiments can be reproduced.

\subsection{Task}

We consider the task of identifying actions performed in videos for two reasons.
First, it has gained much interest in the computer vision community for its many applications in a variety of domains such as intelligent video surveillance, shopping behavioural analysis. 
Second, due to the abundance of smart-phones and social media the amount of videos recorded and uploaded have been increasing and most likely will continue to increase and thus expand the already large video datasets. As the growth rate of videos exceeds the growth rate for other types of data (such as images), tasks on videos are inherently more interesting to tackle.

\subsection{Datasets}
\label{toc:datasets}

In this work we used the following datasets:

\textbf{Kinetics}: approximately 300,000 video clips, covering 400 human action classes \cite{kay2017kinetics}. \\
\textbf{Sports-1M:} 1 million YouTube videos belonging to 487 classes \cite{Karpathy2014}. 
We used roughly 40\% of this dataset as the full size was prohibitive for us to handle.\\
\textbf{UCF101:} 13320 videos from 101 action categories \cite{soomro2012ucf101}. \\
\textbf{HMDB51:} 6,766 video clips extracted from a wide range of sources with 51 distinct action categories \cite{kuehne2011hmdb}.

\subsection{Model Architecture: C3D}

We used the PyTorch implementation\footnote{\url{https://github.com/DavideA/c3d-pytorch}} of C3D \cite{Tran2015}, which was originally implemented in a modified version of BVLC caffe \cite{Jia2014} that supports 3-Dimensional Convolutional Networks \cite{Ji2013}.
This architecture looks similar to popular CNN architectures except that 3D convolutions now replace the 2D convolutions.

\subsection{Training Procedure}
\label{toc:training_procedure}

We test the hypothesis that pre-training on a randomly-labeled dataset will improve learning performance on the \textit{HMDB51} and \textit{UCF-101} datasets.
In order to test the hypothesis, we initially pre-train the network on a randomly-labeled version of one of the datasets, and then use this pre-trained model to train on the correctly labeled version of the other dataset. 
The experimental procedure is as follows:
\begin{enumerate}
    \item Pick a dataset $D$ from  Sports-1M, UCF-101 or HMDB51.
    \item Relabel all training instances of the selected dataset $D$ with randomly chosen labels. 
    \item Train a 3D-CNN on dataset $D$ for $200$ epochs with a inital learning rate of $0.1$, momentum $0.9$ and learning rate decay with a patience of $10$ epochs.
    \item Fine-tune this pre-trained model on the other dataset with the correct labels. 
\end{enumerate}

This procedure is repeated for all the datasets in turn. However, HMDB51 and UCF-101 are cross pre-trained on each other, while Kinetics is pre-trained on the much larger Sports-1M dataset.

%% file: sections/results.tex
\section{Results}
\label{toc:results}

To evaluate the effectiveness of the proposed pre-training method, we initally compare the performance of the C3D network -- with and without pre-training -- on the UCF-101 and HMDB51 datasets. 
However, as both these datasets are fairly small\footnote{In relation to state-of-the-art datasets which are several orders of magnitude bigger}, we also consider pre-training the network on the Sports1M and evaluating it on the Kinetics dataset.
This allows us to measure the impact of scaling up the quantity of the data used for pre-training.
As all the datasets are used for video action recognition, the task is formulated as a classification problem, and as such evaluated using the accuracy metric.

\subsection{HMDB51 and UCF101}

Following the experimental procedure from Section~\ref{toc:training_procedure}, we evaluate the effectiveness on cross pre-training on the HMDB51 and UCF-101 datasets. 
In Table~\ref{table:comparisonsmall} and Figure~\ref{fig:hmdb_ucf}, we can see that the cross pre-training results in a performance gain of $0.8\%$ and $1.7\%$ for HMDB51 and UCF-101 respectively.  


\begin{table}[!t]
\centering
\begin{tabular}{rcc}
\toprule
Method  & UCF-101 & HMDB51\\ \midrule
No pre-training &  $32.9\%$ & $17.4\%$ \\ 
With pre-training & \textbf{34.6\%} & \textbf{18.2\%} \\ 
\bottomrule
\end{tabular}
\caption{Comparison of validation accuracy of C3D --with and without pre-training-- on UCF-101 and HMDB51. 
In this scenario pre-training has been performed on a randomly-labeled version of UCF-101 for HMDB51 and vice-versa for UCF-101.}
\label{table:comparisonsmall}
\vspace{-6mm}
\vspace{6mm}
\vspace{-6mm}
\end{table}

\subsection{Kinetics}

As seen in Table~\ref{table:comparison_kinetics} and Figure~\ref{fig:C3D_kinetics}, the C3D network without pre-training scores $35\%$ and the network pre-trained on the correctly-labeled version of Sports-1M scores $69.8\%$. These allow us to establish a baseline on the Kinetics dataset and an upper bound for what is possible when using transfer learning from the Sports-1M dataset.  
The network that has been pre-trained on the randomly-labeled subset (only $40\%$ of the data is used due to computational constraints) of the Sports-1M dataset scores $40.2\%$, a relative reduction of the error rate by $8\%$. 


\begin{table}[h!]
\centering
\begin{tabular}{r c}
    \toprule
    Pre-training & Accuracy \\ \midrule
    None &  35.0\% \\
    Randomly-labeled Sports-1M & 40.2\% \\
     Correctly-labeled Sports-1M & 69.8\% \\ \bottomrule
\end{tabular}
\caption{Comparison of validation accuracies of the C3D network with: no pre-training, pre-training on the randomly-labeled variant of Sports-1M and pre-training on the correctly-labeled Sports-1M.}
\label{table:comparison_kinetics}
\end{table}

\subsection{Memorization Infrastructure}

As previously mentioned (see Section~\ref{toc:introduction}) previous work suggests that training on random labels forces the network to set up infrastructure for memorization, and that this infrastructure can then be used to make further memorization more efficient \cite{Li2018}.
We verified this hypothesis by training a network for memorization and after a fixed amount of epochs re-shuffled the labels hence forcing the network to start over with the process.
As shown in Figure~\ref{fig:retrain} subsequent memorization processes are much faster than the initial one, thus indicating that the network is in fact building up a memorization infrastructure of some sort.

%% file: sections/conclusion.tex
\section{Conclusion}
\label{toc:conclusion}

In this paper we introduced a novel approach for performing unsupervised pre-training in the form of training for classification with random labels.
Our preliminary experiments suggests that is it possible to learn useful representations by leveraging a large amount of unlabeled data in this way, although the improvement in performances are limited to $1-5$\% of accuracy. 
We believe that further research in this direction could provide larger margin of improvement and allow us to gain a better understanding of deep neural networks.
 
\subsection{Future Work}
\label{toc:future_work}

We plan to further investigate the dynamic of memorization by determining at which stage of the network (early layers close to input or later ones close to final features) are most responsible for it.
We speculate that, even though regularization hinders the memorization capability of the network, it might be 
beneficial to learn more useful feature for a later classification task.
Finally, we want to inspect what the network is looking at in the input to succeed in memorizing every training sample (for example with global average pooling layers \cite{zhou2016learning}).

%% file: main.bbl
\begin{thebibliography}{27}
\providecommand{\natexlab}[1]{#1}
\providecommand{\url}[1]{\texttt{#1}}
\expandafter\ifx\csname urlstyle\endcsname\relax
  \providecommand{\doi}[1]{doi: #1}\else
  \providecommand{\doi}{doi: \begingroup \urlstyle{rm}\Url}\fi

\bibitem[Alberti et~al.(2017)Alberti, Seuret, Ingold, and Liwicki]{alberti2017}
Alberti, M., Seuret, M., Ingold, R., and Liwicki, M.
\newblock {A Pitfall of Unsupervised Pre-Training}.
\newblock In \emph{2017 31st Neural Information Processing Systems (NIPS), Deep
  Learning: Bridging Theory and Practice workshop}, Long Beach, California,
  USA, nov 2017.

\bibitem[Alberti et~al.(2018)Alberti, Pondenkandath, W\"ursch, Bouillon,
  Seuret, Ingold, and Liwicki]{alberti2018tampering}
Alberti, M., Pondenkandath, V., W\"ursch, M., Bouillon, M., Seuret, M., Ingold,
  R., and Liwicki, M.
\newblock {Are You Tampering With My Data?}
\newblock In \emph{2018 15th European Conference on Computer Vision (ECCV),
  Workshop on Objectionable Content and Misinformation (WOCM)}, Munich,
  Germany, sep 2018.

\bibitem[Baldi(2012)]{Baldi2012}
Baldi, P.
\newblock {Autoencoders, Unsupervised Learning, and Deep Architectures}.
\newblock \emph{ICML Unsupervised and Transfer Learning}, 2012.
\newblock ISSN 0899-7667.
\newblock \doi{10.1561/2200000006}.

\bibitem[Bengio et~al.(2007)Bengio, Lamblin, Popovici, and
  Larochelle]{Bengio2007}
Bengio, Y., Lamblin, P., Popovici, D., and Larochelle, H.
\newblock {Greedy Layer-Wise Training of Deep Networks}.
\newblock \emph{Advances in Neural Information Processing Systems}, 2007.
\newblock ISSN 01628828.
\newblock \doi{citeulike-article-id:4640046}.

\bibitem[Bengio et~al.(2013)Bengio, Courville, and Vincent]{Bengio2013}
Bengio, Y., Courville, A., and Vincent, P.
\newblock {Representation learning: A review and new perspectives}.
\newblock \emph{IEEE Transactions on Pattern Analysis and Machine
  Intelligence}, 2013.
\newblock ISSN 01628828.
\newblock \doi{10.1109/TPAMI.2013.50}.

\bibitem[Dosovitskiy et~al.(2014)Dosovitskiy, Springenberg, Riedmiller, and
  Brox]{Dosovitskiy2014}
Dosovitskiy, A., Springenberg, J.~T., Riedmiller, M., and Brox, T.
\newblock {Discriminative Unsupervised Feature Learning with Convolutional
  Neural Networks}.
\newblock \emph{Advances in Neural Information Processing Systems 27}, pp.\
  766--774, 2014.
\newblock ISSN 0162-8828.
\newblock \doi{10.1109/TPAMI.2015.2496141}.

\bibitem[Dundar et~al.(2015)Dundar, Jin, and
  Culurciello]{dundar2015convolutional}
Dundar, A., Jin, J., and Culurciello, E.
\newblock Convolutional clustering for unsupervised learning.
\newblock \emph{arXiv preprint arXiv:1511.06241}, 2015.

\bibitem[Erhan et~al.(2009)Erhan, Manzagol, Bengio, Bengio, and
  Vincent]{Erhan2009}
Erhan, D., Manzagol, P.-A., Bengio, Y., Bengio, S., and Vincent, P.
\newblock The difficulty of training deep architectures and the effect of
  unsupervised pre-training.
\newblock In \emph{Artificial Intelligence and Statistics}, pp.\  153--160,
  2009.

\bibitem[Erhan et~al.(2010)Erhan, Bengio, Courville, Manzagol, Vincent, and
  Bengio]{erhan2010does}
Erhan, D., Bengio, Y., Courville, A., Manzagol, P.-A., Vincent, P., and Bengio,
  S.
\newblock Why does unsupervised pre-training help deep learning?
\newblock \emph{Journal of Machine Learning Research}, 11\penalty0
  (Feb):\penalty0 625--660, 2010.

\bibitem[Glorot \& Bengio(2010)Glorot and Bengio]{Glorot2010}
Glorot, X. and Bengio, Y.
\newblock {Understanding the difficulty of training deep feedforward neural
  networks}.
\newblock \emph{PMLR}, 9:\penalty0 249--256, 2010.
\newblock ISSN 15324435.
\newblock \doi{10.1.1.207.2059}.

\bibitem[Hinton et~al.(2006)Hinton, Osindero, and Teh]{hinton2006fast}
Hinton, G.~E., Osindero, S., and Teh, Y.-W.
\newblock A fast learning algorithm for deep belief nets.
\newblock \emph{Neural computation}, 18\penalty0 (7):\penalty0 1527--1554,
  2006.

\bibitem[Jain(2010)]{Jain2010}
Jain, A.~K.
\newblock {Data clustering: 50 years beyond K-means}.
\newblock \emph{Pattern Recognition Letters}, 2010.
\newblock ISSN 01678655.
\newblock \doi{10.1016/j.patrec.2009.09.011}.

\bibitem[Ji et~al.(2013)Ji, Xu, Yang, and Yu]{Ji2013}
Ji, S., Xu, W., Yang, M., and Yu, K.
\newblock {3D Convolutional neural networks for human action recognition}.
\newblock \emph{IEEE Transactions on Pattern Analysis and Machine
  Intelligence}, 35\penalty0 (1):\penalty0 221--231, jan 2013.
\newblock ISSN 01628828.
\newblock \doi{10.1109/TPAMI.2012.59}.

\bibitem[Jia et~al.(2014)Jia, Shelhamer, Donahue, Karayev, Long, Girshick,
  Guadarrama, and Darrell]{Jia2014}
Jia, Y., Shelhamer, E., Donahue, J., Karayev, S., Long, J., Girshick, R.,
  Guadarrama, S., and Darrell, T.
\newblock {Caffe: Convolutional Architecture for Fast Feature Embedding}.
\newblock \emph{dl.acm.org}, 2014.
\newblock ISSN 10636919.
\newblock \doi{10.1145/2647868.2654889}.

\bibitem[Karpathy et~al.(2014)Karpathy, Toderici, Shetty, Leung, Sukthankar,
  and Li]{Karpathy2014}
Karpathy, A., Toderici, G., Shetty, S., Leung, T., Sukthankar, R., and Li,
  F.~F.
\newblock {Large-scale video classification with convolutional neural
  networks}.
\newblock \emph{Proc. IEEE CVPR}, 2014.
\newblock ISSN 10636919.
\newblock \doi{10.1109/CVPR.2014.223}.

\bibitem[Kay et~al.(2017)Kay, Carreira, Simonyan, Zhang, Hillier,
  Vijayanarasimhan, Viola, Green, Back, Natsev, et~al.]{kay2017kinetics}
Kay, W., Carreira, J., Simonyan, K., Zhang, B., Hillier, C., Vijayanarasimhan,
  S., Viola, F., Green, T., Back, T., Natsev, P., et~al.
\newblock The kinetics human action video dataset.
\newblock \emph{arXiv preprint arXiv:1705.06950}, 2017.

\bibitem[Kuehne et~al.(2011)Kuehne, Jhuang, Garrote, Poggio, and
  Serre]{kuehne2011hmdb}
Kuehne, H., Jhuang, H., Garrote, E., Poggio, T., and Serre, T.
\newblock Hmdb: a large video database for human motion recognition.
\newblock In \emph{Computer Vision (ICCV), 2011 IEEE International Conference
  on}, pp.\  2556--2563. IEEE, 2011.

\bibitem[Li et~al.(2018)Li, Farkhoor, Liu, and Yosinski]{Li2018}
Li, C., Farkhoor, H., Liu, R., and Yosinski, J.
\newblock {Measuring the Intrinsic Dimension of Objective Landscapes}.
\newblock \emph{arxiv.org}, apr 2018.

\bibitem[Masci et~al.(2011)Masci, Meier, Cireşan, and Schmidhuber]{Masci2011}
Masci, J., Meier, U., Cireşan, D., and Schmidhuber, J.
\newblock {Stacked convolutional auto-encoders for hierarchical feature
  extraction}.
\newblock In \emph{Lecture Notes in Computer Science (including subseries
  Lecture Notes in Artificial Intelligence and Lecture Notes in
  Bioinformatics)}, 2011.
\newblock ISBN 9783642217340.
\newblock \doi{10.1007/978-3-642-21735-7_7}.

\bibitem[Radford et~al.(2015)Radford, Metz, and Chintala]{Radford2015}
Radford, A., Metz, L., and Chintala, S.
\newblock {Unsupervised Representation Learning with Deep Convolutional
  Generative Adversarial Networks}.
\newblock \emph{arXiv}, 2015.
\newblock ISSN 0004-6361.
\newblock \doi{10.1051/0004-6361/201527329}.

\bibitem[Soomro et~al.(2012)Soomro, Zamir, and Shah]{soomro2012ucf101}
Soomro, K., Zamir, A.~R., and Shah, M.
\newblock Ucf101: A dataset of 101 human actions classes from videos in the
  wild.
\newblock \emph{arXiv preprint arXiv:1212.0402}, 2012.

\bibitem[Statistics(2014)]{youtubestats}
Statistics, Y.
\newblock Statistics - youtube, 2014.
\newblock URL
  \url{https://web.archive.org/web/20141213110649/https://www.youtube.com/yt/press/statistics.html}.

\bibitem[Tran et~al.(2015)Tran, Bourdev, Fergus, Torresani, and
  Paluri]{Tran2015}
Tran, D., Bourdev, L., Fergus, R., Torresani, L., and Paluri, M.
\newblock {Learning Spatiotemporal Features with 3D Convolutional Networks}.
\newblock In \emph{2015 IEEE International Conference on Computer Vision
  (ICCV)}, 2015.
\newblock ISBN 978-1-4673-8391-2.
\newblock \doi{10.1109/ICCV.2015.510}.

\bibitem[Yang et~al.(2015)Yang, Wang, Lin, Wipf, Guo, and Guo]{Yang2015}
Yang, H., Wang, B., Lin, S., Wipf, D., Guo, M., and Guo, B.
\newblock {Unsupervised extraction of video highlights via robust recurrent
  auto-encoders}.
\newblock In \emph{Proceedings of the IEEE International Conference on Computer
  Vision}, 2015.
\newblock ISBN 9781467383912.
\newblock \doi{10.1109/ICCV.2015.526}.

\bibitem[Zhang et~al.(2016)Zhang, Bengio, Hardt, Recht, and Vinyals]{Zhang2016}
Zhang, C., Bengio, S., Hardt, M., Recht, B., and Vinyals, O.
\newblock {Understanding deep learning requires rethinking generalization}.
\newblock \emph{arxiv.org}, 2016.
\newblock ISSN 10414347.
\newblock \doi{10.1109/TKDE.2015.2507132}.

\bibitem[Zhao et~al.(2015)Zhao, Mathieu, Goroshin, and LeCun]{Zhao2015}
Zhao, J., Mathieu, M., Goroshin, R., and LeCun, Y.
\newblock {Stacked What-Where Auto-encoders}.
\newblock \emph{arxiv.org}, jun 2015.

\bibitem[Zhou et~al.(2016)Zhou, Khosla, Lapedriza, Oliva, and
  Torralba]{zhou2016learning}
Zhou, B., Khosla, A., Lapedriza, A., Oliva, A., and Torralba, A.
\newblock Learning deep features for discriminative localization.
\newblock In \emph{Proceedings of the IEEE Conference on Computer Vision and
  Pattern Recognition}, pp.\  2921--2929, 2016.

\end{thebibliography}
